\documentclass[letterpaper, 10 pt, conference]{ieeeconf}
\usepackage{subfigure}
\usepackage[utf8]{inputenc}
\usepackage[dvipdfmx]{graphicx}
\usepackage{amsmath}
\usepackage{url}
\usepackage{color}
\usepackage{cite}
\usepackage{algorithm}
\usepackage{algorithmic}
\usepackage{amssymb}
\usepackage{amsfonts}
\usepackage{multirow}
\usepackage{booktabs}

\IEEEoverridecommandlockouts
\title{\LARGE \bf
Real-time Multi-Plane Segmentation Based on GPU Accelerated High-Resolution 3D Voxel Mapping for Legged Robot Locomotion
}

\author{Shun Niijima$^{1}$, Ryoichi Tsuzaki$^{1}$, Noriaki Takasugi$^{1}$ and Masaya Kinoshita$^{1}$
\thanks{$^{1}$ Authors are with sony group corporation, Minato-ku, Tokyo, Japan, 108-0075
        {\tt\small Shun.Niijima@sony.com}}%
}

\begin{document}
\maketitle
\thispagestyle{empty}
\pagestyle{empty}

\begin{abstract}
  This paper proposes a real-time multi-plane segmentation method based on GPU-accelerated high-resolution 3D voxel mapping for legged robot locomotion. Existing online planar mapping approaches struggle to balance accuracy and computational efficiency: direct depth image segmentation from specific sensors suffers from poor temporal integration, height map-based methods cannot represent complex 3D structures like overhangs, and voxel-based plane segmentation remains unexplored for real-time applications. To address these limitations, we develop a novel framework that integrates vertex-based connected component labeling with random sample consensus based plane detection and convex hull, leveraging GPU parallel computing to rapidly extract planar regions from point clouds accumulated in high-resolution 3D voxel maps. Experimental results demonstrate that the proposed method achieves fast and accurate 3D multi-plane segmentation at over 30 Hz update rate even at a resolution of 0.01 m, enabling the detected planes to be utilized in real time for locomotion tasks. Furthermore, we validate the effectiveness of our approach through experiments in both simulated environments and physical legged robot platforms, confirming robust locomotion performance when considering 3D planar structures.
\end{abstract}

\section{INTRODUCTION}
\begin{figure}[t]
  \centering
  \includegraphics[width=0.65\columnwidth]{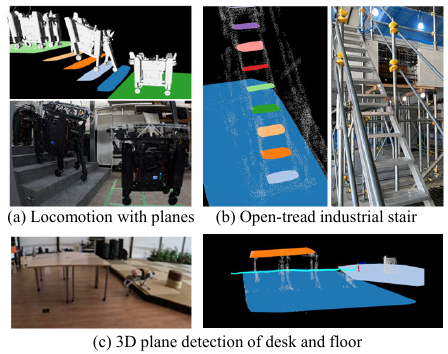}
  \caption{
    The proposed method enables real-time high-precision 3D multi-plane segmentation for legged robot locomotion (a), detects 3D multi-layered planar surfaces (b), and enables 3D locomotion under structures such as desks (c).
  }
  \label{fig:proposed_demo}
\end{figure}

Legged robots enable traversal of challenging 3D structures by utilizing discrete footholds, facilitating locomotion in environments inaccessible to wheeled platforms, as illustrated in Fig.~\ref{fig:proposed_demo}(a). This versatility makes them promising for diverse applications such as exploration, surveillance, and autonomous inspection~\cite{bhatti2015survey,yoshiike2017development,gehring2021anymal,chen2024autonomous}.

To achieve safe and efficient locomotion, legged robots require accurate recognition of traversable foothold regions and rapid detection of stable planar surfaces\cite{grandia2021multi,sato2022robust,perceptive_locomotion2023}. For this purpose, height maps have been widely utilized~\cite{elevation2018,elevation_gpu2022,mem2023}, representing environments as 2.5-dimensional structures with a single height value for each $(x,y)$ coordinate. While computationally efficient, height maps fundamentally fail to represent multiple planes at identical coordinates, making it challenging to model multi-layered surfaces and overhanging structures. This limitation results in collision risks and locomotion failures when traversing open-tread stairs or locomotion beneath structures such as tables, as illustrated in Fig.\ref{fig:proposed_demo}(b)(c).

To address these challenges, several studies have explored extensions to 3D voxel mapping for locomotion\cite{supervoxel_plane2017,bertrand2020detecting}.  However, existing frameworks struggle to balance computational speed and accuracy due to the increased processing time caused by the management of 3D voxels and the substantial increase in the number of point clouds. This computational overhead severely constrains the high-speed locomotion capability of legged robots. Although some approaches attempt to ensure real-time performance by lowering the resolution, such measures inevitably sacrifice detection accuracy and environmental adaptability. Therefore, existing frameworks still fail to achieve real-time, high-precision 3D multi-plane detection.

This research addresses this challenge by proposing a real-time multi-plane segmentation framework based on GPU-accelerated high-resolution 3D voxel mapping. The proposed framework combines a 3D voxel mapping module with a GPU-accelerated multi-plane segmentation using connected component labeling (CCL) clustering and cluster based parallelized plane boundary estimation to enable rapid extraction of planar regions and their boundaries from large-scale point clouds. The framework achieves real-time multi-plane segmentation in the robot's vicinity while maintaining an exceptionally high resolution of 0.01 m within the 3D voxel representation.

The primary contributions of this research are as follows:
\begin{itemize}
  \item Proposal of a GPU-accelerated 3D multi-plane segmentation utilizing CCL clustering and cluster based parallelized plane boundary estimation method.
  \item Implementation of a comprehensive framework combining 3D voxel mapping with multi-plane segmentation to enable environmental perception for legged robot locomotion.
  \item Comprehensive experimental validation demonstrating the effectiveness and successful deployment of the proposed method on legged robot platforms for safe 3D locomotion tasks.
\end{itemize}

\section{RELATED WORK}
\subsection{Environmental perception for legged robot locomotion}
Reliable plane segmentation and polygonization methods are widely adopted as environmental representation techniques for legged robot locomotion tasks~\cite{deits2014,tonneau2018,perceptive_locomotion2023}. These representations provide geometrically accurate and computationally efficient environmental information, enabling fast computation in model-based legged robot locomotion control systems. Unlike wheeled mobile robots, which typically require detection of a single dominant plane, legged robots must identify and segment multiple planar regions in 3D space such as stairs and stepping stones to plan discrete foothold placements and ensure stable locomotion~\cite{grilli2017review,sun2024review}.

Recent advances in reinforcement learning (RL) have enabled legged robots to achieve robust locomotion over challenging terrain~\cite{miki2022learning,xu2024dexterous}. To enhance safety, researchers have explored combining RL with model-based approaches~\cite{kovalev2023combining,yang2023continuous,yang2024agile}. Furthermore, plane segmentation plays a crucial role in lightweight real-to-sim frameworks by providing geometrically accurate and computationally efficient environment reconstruction, particularly for indoor settings~\cite{tan2025planarsplatting}, enabling precise environmental mapping for safe locomotion. Regardless of the control paradigm, high-resolution environmental mapping and accurate planar region estimation remain one of important tasks for safe locomotion.

\subsection{Multi-plane segmentation for locomotion}
Two primary approaches have been proposed to realize the aforementioned plane segmentation capabilities. 

\subsubsection*{(i) Direct segmentation from sensor data}
The first approach directly estimates planar regions from sensor data\cite{polytopic2022, planar_gpu2021, polygon_mapping2024, polylidar3d2020}. These methods can rapidly extract planar regions on a per-frame basis and support dynamic locomotion of legged robots. However, this approach faces two main challenges: (1) dependence on dense depth images and aligned point clouds, making it unsuitable for sparse LiDAR observations and non-repetitive scanning patterns \cite{liu2021low}, and (2) reliance on heuristic temporal integration of planar regions across frames, which often leads to inaccurate correspondence and tracking in dynamic environments. Although some methods incorporate GPU-accelerated multi-plane detection, the aforementioned constraints leave challenges in representing 3D environments under diverse sensing configurations and dynamic changes.

\subsubsection*{(ii) Map-based plane estimation}
The second approach estimates planar regions on robot-centric maps by accumulating temporal sensor information. Height map-based methods \cite{grandia2021multi,perceptive_locomotion2023} have been extensively utilized.
In particular, the methods proposed in \cite{perceptive_locomotion2023,elevation_gpu2022} is a widely used height map-based framework that achieves rapid ray casting for handling dynamic objects and efficient traversability estimation by performing height map updates directly on the GPU. Nevertheless, multi-plane segmentation is executed using the height map transferred to the CPU, which presents opportunities for improvement in terms of processing efficiency.
By applying GPU-accelerated image-based multi-plane segmentation methods, such as those described in \cite{planar_gpu2021, polygon_mapping2024}, for height maps generated on the GPU, it may be possible to improve processing time.  
However, approaches that convert 3D space into 2D representations, such as height maps, have inherent limitations in reconstruction 3D environments including tunnels, overhangs, and multi-layered surfaces. Consequently, upper planes and structural elements are omitted, which increases the risk of collisions.

To address these limitations, several studies have explored 3D point cloud/voxel framework including multi-plane segmentation for locomotion~\cite{supervoxel_plane2017,bertrand2020detecting,t3_perceptive2024}. While these methods provide richer 3D representations, they operate at low update rates of 1--2~Hz due to CPU-based processing, failing to fully leverage modern sensors capable of acquiring data at 10~Hz or higher frequencies.

\subsection{GPU-accelerated 3D voxel mapping}
Recent studies have leveraged GPU acceleration for the efficient construction of high-resolution 3D voxel maps and occupancy map~\cite{nvblox2024,hoss2024covoxslam,overbye2022g,stepanas2022ohm}.  Compared to point cloud-based methods, voxel-based representations regularize the spatial structure, enabling highly efficient parallel execution on GPUs by avoiding the costly and irregular memory access patterns associated with nearest neighbor searches, such as kd-tree and octree.  These approaches facilitate real-time generation of occupancy grids and Euclidean/Truncated Signed Distance Fields (ESDF/TSDF), and have been widely adopted for navigation tasks such as path planning and simultaneous localization and mappin (SLAM) applications.

However, these framework do not incorporate multi-plane segmentation capabilities and therefore cannot directly support plane-based locomotion control. Notably, the integration of high-resolution 3D planar information obtained in real-time from 3D maps into legged robot decision-making processes remains an important yet insufficiently explored research area. 

The proposed framework addresses this gap by accumulating point clouds in 3D voxel maps on GPU and integrating GPU-accelerated multi-plane segmentation modules to generate polygonal plane representations. It employs methods well-suited for GPU parallelization, including vertex-based CCL clustering and multi-cluster plane estimation, and is designed to flexibly support diverse sensor configurations for broad applicability across robotic platforms. We demonstrate real-time multi-plane segmentation around the robot ($5.0,\mathrm{m}\times5.0,\mathrm{m}\times5.0,\mathrm{m}$) at 0.01 m resolution, and validate its effectiveness through applications in 3D locomotion tasks.

\begin{figure*}[!h]
  \centering
  \includegraphics[width=1.5\columnwidth]{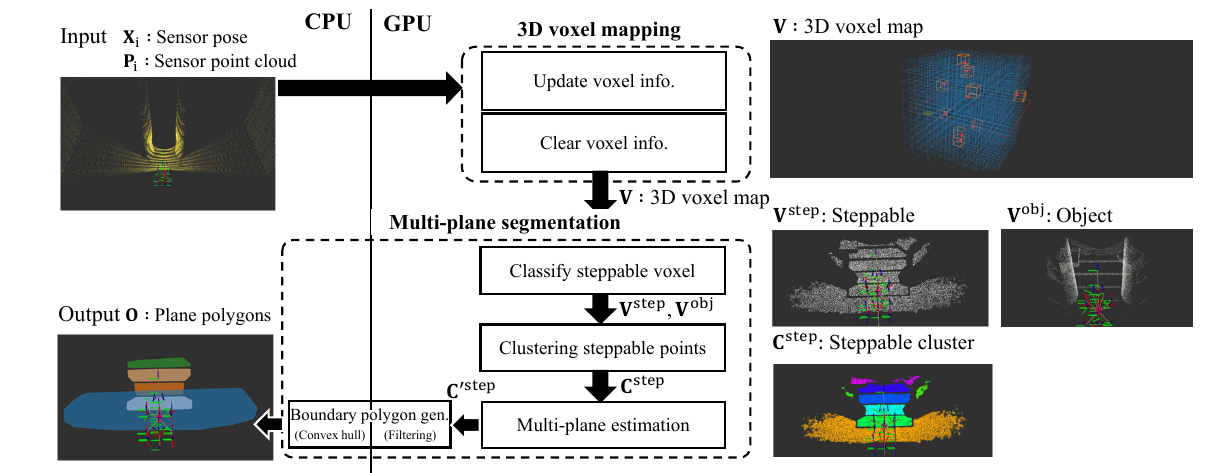}
  \caption{
    Framework overview of the proposed method. The framework consists of a mapping module and a multi-plane segmentation module. The mapping module accumulates point clouds in a 3D voxel map and removes dynamic objects through ray casting operations. The multi-plane segmentation module classifies accumulated points into steppable and object points, clusters the steppable points, and detects multiple planes from the clustered steppable points.
  }
  \label{fig:system_overview}
\end{figure*}

\section{METHODS}
\subsection{Framework overview}
Fig.~\ref{fig:system_overview} illustrates the overall architecture of the proposed framework, which comprises a 3D voxel mapping module and a multi-plane segmentation module. The 3D voxel mapping module receives sensor point clouds ${\mathbf P}_i = \{x, y, z\}$ and robot pose ${\mathbf X}_i \in SE(3)$ as inputs, and efficiently accumulates point clouds into the voxel map ${\mathbf V}$. Section III.B details our streamlined implementation of the 3D voxel mapping approach.

The multi-plane segmentation module estimates stable planar regions from accumulated point clouds in ${\mathbf V}$ for legged robot locomotion tasks and outputs these regions as polygons ${\mathbf O}$. Initially, the module classifies voxels into steppable voxels ${\mathbf V}^{\text{step}}$ and object voxels ${\mathbf V}^{\text{obj}}$ based on normal vector directions computed from surrounding voxel point clouds. Subsequently, steppable points are clustered to generate multiple steppable clusters ${\mathbf C}^{\text{step}}$. Finally, plane parameters and regions are calculated for each steppable cluster and converted to polygonal representations ${\mathbf O}$ for output. 
Section III.C presents the proposed multi-plane segmentation methodology.

\subsection{3D voxel mapping}
\subsubsection{3D voxel map representation}
\begin{figure}[h]
  \centering
  \includegraphics[width=0.5\columnwidth]{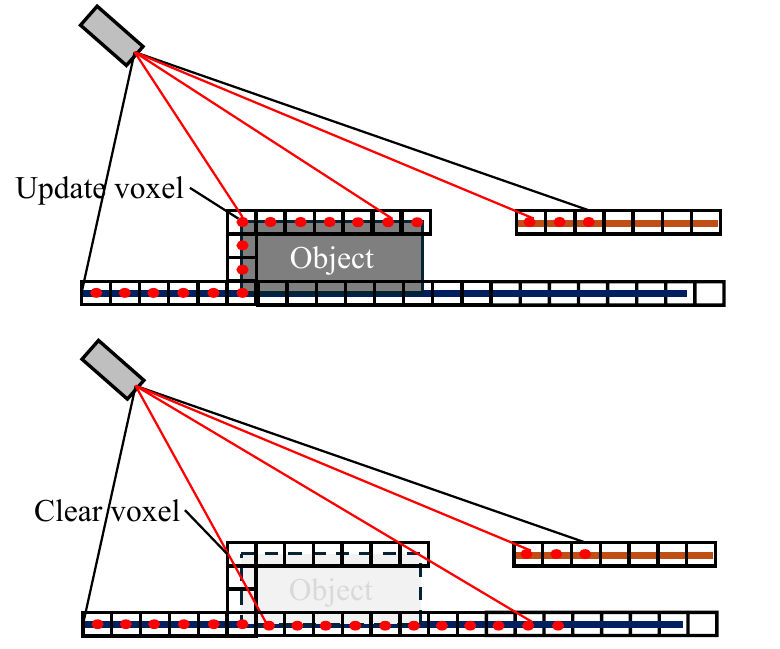}
  \caption{
    3D voxel mapping module: The voxel update module accumulates point clouds in the 3D voxel map, while the voxel clearing module removes dynamic objects by casting rays from the sensor to each point cloud location.
    }
  \label{fig:ray_casting}
\end{figure}

Each voxel in the 3D voxel map maintains average position and point count information and status ${\mathbf v}_i = \{{\mathbf \mu}_i, c_i, s_i\}$, where ${\mathbf \mu}_i$ represents the voxel's average position, $c_i$ denotes the number of points within the voxel, and $s_i$ indicates the voxel status. The status is represented by three discrete values: 0 (free space), 1 (occupied by object), and 2 (steppable surface). The center of the 3D voxel map dynamically translates to maintain alignment with the robot-centric coordinate frame. 
Fig. \ref{fig:ray_casting} illustrates the two sub-modules: voxel information update and voxel information clearing for dynamic object removal.  In the mapping module, threads are allocated corresponding to the number of sensor point clouds, executing parallel operations.

\subsubsection{Voxel information update}
During map updates, each thread performs parallel coordinate transformation of point clouds to the robot-centric coordinate system and updates voxel average position information. For each point ${\mathbf p}_i$ transformed to the robot-centric coordinate system, the update of average position ${\mathbf \mu}_i$, point count $c_i$, and status $s_i$ of the corresponding voxel ${\mathbf v}_i$ is formulated as follows:
\begin{equation}
  {\mathbf v}_i = \left\{
    \begin{array}{ll}
      \{{\mathbf p}_i, 1, \text{occupied}\}  & \text{if } c_i = 0, \\
      \{\frac{1}{c_i + 1} \left( c_i {\mathbf \mu}_i + {\mathbf p}_i \right), c_i + 1, \text{occupied}\} & \text{otherwise}.
    \end{array}
  \right.
\end{equation}

\subsubsection{Voxel information clearing}
The voxel clearing module utilizes sensor pose $\mathbf X_{i}$ and point cloud $\mathbf P_i$ information to enable adaptation to dynamic environmental changes. Ray trajectories from the sensor to each point cloud are computed, and point clouds existing in voxels along those ray paths are identified and removed as follows: 
\begin{equation}
    \begin{array}{ll}
  {\mathbf v}_i = \{None, 0, \text{free}\} & \text{if } n_i > 0.
    \end{array}
\end{equation}
For implementation efficiency, voxel indices targeted for removal are precomputed, and duplicate indices are eliminated to prevent concurrent access from multiple threads. This optimization enables high-speed dynamic object removal.

\subsection{Multi-plane segmentation}
\subsubsection{Steppable point classification}
The steppable point classification module categorizes the accumulated 3D voxel map $\mathbf{V}$ into steppable points $\mathbf{V}^{\text{step}}$ and object points $\mathbf{V}^{\text{obj}}$. To identify steppable surfaces efficiently, this module evaluates the orientation of local normal vectors and the density of neighboring points. 

Each occupied voxel collects points ${\mathbf P}_j$ from neighboring voxel ${\mathbf V}_j$ that are specified voxel range centered on the target voxel.
The normal vector of each voxel is estimated by computing the covariance matrix $\boldsymbol{\Sigma}_i$ from its neighboring points $\mathbf{P}_j$. Eigen-decomposition of $\boldsymbol{\Sigma}_i$ is performed using the Jacobi method~\cite{golub2013matrix}, and the normal vector ${\mathbf n}_i$ is obtained from the eigenvector corresponding to the smallest eigenvalue:
\begin{equation}
  \label{eq:eigen}
  \mathbf{n}_i = \text{eigenvector}(\boldsymbol{\Sigma}_i)_{\lambda_{\text{min}}}.
\end{equation}

Classification into steppable points is then performed based on the angle $\theta_i$ between $\mathbf{n}_i$ and the gravity vector, and on the number of neighbor points $N$. A point is considered steppable if both conditions are satisfied:
\begin{equation}
  \label{eq:classify}
  \text{Steppability} =
  \begin{cases}
    1 & \text{if } N \geq N_{\text{th}} \text{ and } \theta_i \leq \theta_{\text{th}}, \\
    0 & \text{otherwise}.
  \end{cases}
\end{equation}
Here, $N_{\text{th}}$ and $\theta_{\text{th}}$ represent the thresholds for neighbor count and normal angle, respectively, 
which are set to $N_{\text{th}}=3$ and $\theta_{\text{th}}=15^\circ$ in our implementation.
Voxels with steppability $=1$ are classified as steppable $\mathbf{V}^{\text{step}}$, while the others are categorized as object $\mathbf{V}^{\text{obj}}$. All computations are parallelized by assigning individual threads to occupied voxels.

\subsubsection{Clustering of steppable points}
For clustering steppable points, we implemented a vertex-based CCL method that utilizes both distance and normal angle criteria. As shown in Algorithm~\ref{alg:clustering}, for each steppable voxel $\mathbf{v}_i$, the representative point $\mathbf{\mu}_i$ and those of its neighboring voxels $\mathbf{\mu}_j$ are compared; if the distance is within $d_{\text{th}}$ and the normal angle difference is less than $\theta_{\text{th}}$, the neighbor $\mathbf{v}_j$ is added to the adjacency set $N(\mathbf{v}_i)$ (lines~1--7).

Subsequently, CCL performs parallel label propagation to unify points belonging to the same cluster (lines~8--22). Each thread compares its own label $L_i$ with the labels $L_j$ of its neighbors, and if they differ, both are updated to the smaller label value. This process is iterated until no label changes occur, with atomic operations used to control update conflicts.
Finally, point clouds sharing the same label are grouped as clusters ${\mathbf C}^{\text step}$ and passed to subsequent processing stages.

For further efficiency in GPU implementation, we introduce hierarchical termination detection as in \cite{gpu_euclidean2020}. Specifically, a local flag $m_l$ detects updates within each thread, a block-level flag $m_b$ aggregates updates within each block, and a representative thread updates the global flag $m_g$ to determine overall convergence. This hierarchical structure reduces global memory access and enables highly efficient clustering.

\begin{algorithm}[h]
  \caption{GPU-based Clustering of Steppable Points}
  \label{alg:clustering}
  \begin{algorithmic}[1]
    \REQUIRE Steppable voxels $\mathbf{V}^{\text{step}} = \{\mathbf{v}_i\}$, average positions $\{\mathbf{\mu}_i\}$, normals $\{\mathbf{n}_i\}$, distance threshold $d_{\text{th}}$, normal angle threshold $\theta_{\text{th}}$
    \ENSURE Cluster labels $\{L_i\}$ for all $\mathbf{v}_i \in \mathbf{V}^{\text{step}}$
    \FORALL{voxel $\mathbf{v}_i \in \mathbf{V}^{\text{step}}$ in parallel}
      \FORALL{neighbor voxel $\mathbf{v}_j$ of $\mathbf{v}_i$}
        \IF{$\|\mathbf{\mu}_i - \mathbf{\mu}_j\| < d_{\text{th}}$ \AND $\angle(\mathbf{n}_i, \mathbf{n}_j) < \theta_{\text{th}}$}
          \STATE Add $\mathbf{v}_j$ to adjacency set $\mathcal{N}(\mathbf{v}_i)$
        \ENDIF
      \ENDFOR
    \ENDFOR

    \STATE Initialize labels: $L_i \gets i$ for all $i$
    \STATE Global flag $m_g \gets \text{true}$
    \WHILE{$m_g = \text{true}$}
      \STATE $m_g \gets \text{false}$
      \FORALL{voxel $\mathbf{v}_i \in \mathbf{V}^{\text{step}}$ in parallel}
        \FORALL{$\mathbf{v}_j \in \mathcal{N}(\mathbf{v}_i)$}
          \IF{$L_i > L_j$}
            \STATE $\text{atomic}(L_i \gets L_j)$
          \ELSIF{$L_i < L_j$}
            \STATE $\text{atomic}(L_j \gets L_i)$
          \ENDIF
        \ENDFOR
      \ENDFOR
      \STATE Hierarchical termination detection ($m_l, m_b, m_g$)
    \ENDWHILE
    \RETURN Cluster labels $\{L_i\}$
  \end{algorithmic}
\end{algorithm}

\subsubsection{Multi-plane estimation}
To efficiently process point cloud data, we implemented a cluster-parallel RANSAC method capable of handling multiple clusters simultaneously. This approach extends RANSAC by enabling independent parallelization not only over the number of samples and iterations, but also across multiple clusters, thereby significantly improving computational efficiency for multi-plane estimation.

The cluster-parallel RANSAC procedure consists of the following three stages:
\begin{itemize}
\item[a)] \textbf{Plane parameter estimation}: For each cluster ${\mathbf C}^{\text{step}}_i$, a specified number of RANSAC iterations are performed by randomly selecting three points to estimate plane parameters ${\mathbf \Pi}_i$. Parallelization is achieved across both clusters and iteration nums $I = 100$, with each thread maintaining an independent random state.
\item[b)] \textbf{Inlier counting}: For each estimated plane parameter, the distance to all points within the corresponding cluster is computed, and points within a threshold $\epsilon = 0.01m$ are counted as inliers. This step is parallelized over clusters, points, and iterations.
\item[c)] \textbf{Optimal parameter selection}: For each cluster ${\mathbf C}^{\text{step}}_i$, the plane parameters with the maximum number of inliers are selected parallelized over cluster, and the corresponding inlier points ${\mathbf C}^{'\text{step}}_i$ are extracted parallelization across clusters and inlier points.
\end{itemize}

\subsubsection{Boundary polygon generation}
For boundary polygon generation, we employ a GPU-adapted convex hull algorithm~\cite{gpu_convex_hull_2022}. This method efficiently filters the majority of points on the GPU, while the final convex hull (polygonization) step, which requires sequential processing, is executed on the CPU using only a minimal set of candidate points transferred from the GPU. In this study, we extend this approach to support cluster-level parallelization for enhanced efficiency.

\section{Experiments}
This section presents a comprehensive evaluation of the proposed framework. We first assess computational performance and plane detection accuracy using high-resolution voxel maps in simulation. We then evaluate the impact of cluster-level parallelization on processing speed. Finally, we demonstrate the method's practical utility for legged robot locomotion in both simulated and real-world experiments.
\subsection{Experimental Settings}
\begin{figure}[h]
  \centering
  \begin{tabular}{c}
    \includegraphics[width=0.55\columnwidth]{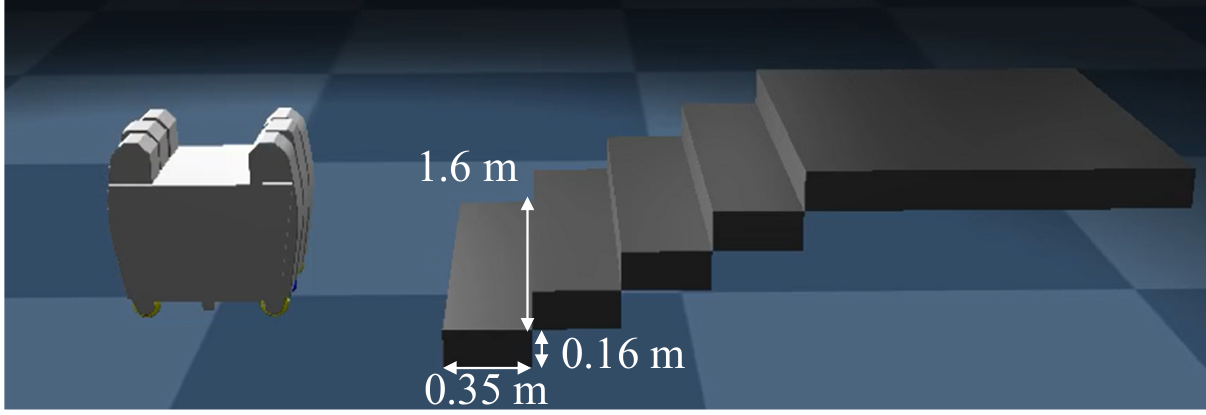} \\
    (a) Five-step stair \\
    \includegraphics[width=0.55\columnwidth]{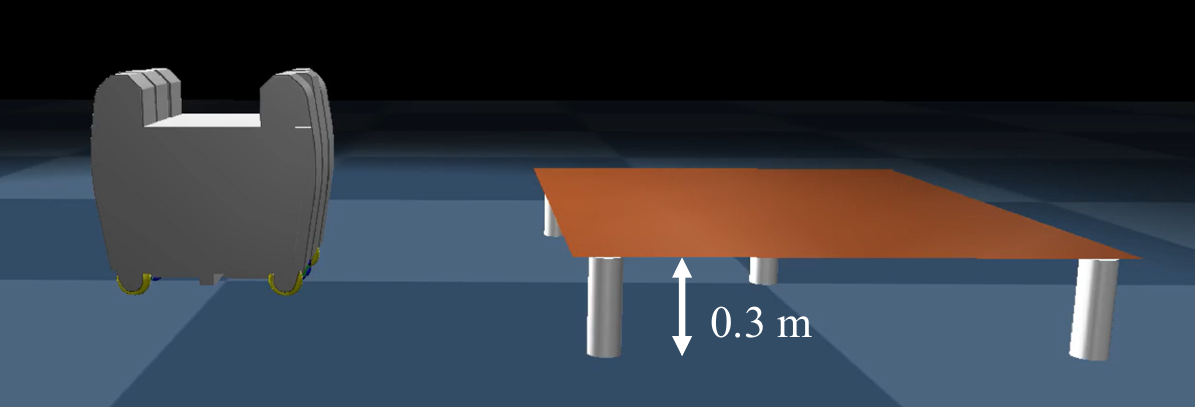} \\
    (b) Single stage (w.o. side plane) \\
  \end{tabular}
  \caption{
    Simulation environments used for evaluation.
  }
  \label{fig:sim_env}
\end{figure}

Fig.~\ref{fig:sim_env} shows two environments: (a) a five-step stair and (b) a single-stage platform. 
We generated sensor data by navigating a legged robot through these scenes in MuJoCo~\cite{todorov2012mujoco}. 
The robot-centric mapping volume was fixed to $5.0\,\mathrm{m}\times5.0\,\mathrm{m}\times5.0\,\mathrm{m}$ with a voxel size of $0.01\ \mathrm{m}$.
We simulated a depth camera (Intel RealSense D435 with 720×480 pixel resolution
) and LiDARs (
  Livox Mid--70,
  Livox Mid--360,
  RoboSense Airy
  ) with their respective scan patterns.
All modules of the proposed framework were executed on a Jetson AGX Orin~\footnote{\url{https://www.nvidia.com/en-us/autonomous-machines/embedded-systems/jetson-agx-orin/}} equipped with an 8-core ARM CPU, a 2048-core Ampere GPU, and 64\,GB of memory.  Unless otherwise specified, plane separation thresholds were set to distance $d_{\text{th}}{=}0.05\,\mathrm{m}$.

To evaluate the effectiveness of the proposed comprehensive GPU framework based on 3D voxel mapping and the proposed GPU-accelerated multi-plane segmentation (MPS) method, we conduct performance comparisons using the following frameworks:
\begin{enumerate}
  \item \textbf{Height map-based(GPU height map+CPU MPS)}\cite{perceptive_locomotion2023}: State-of-the-art height map-based framework utilizing GPU-based elevation map generation and CPU-based multi-plane segmentation.
  \item \textbf{GPU-CPU hybrid framework(GPU 3D voxel mapping+CPU MPS)}: Method combining the proposed GPU-based 3D voxel mapping with conventional CPU-based multi-plane segmentation used in~\cite{t3_perceptive2024}.
  \item \textbf{Proposed framework(GPU 3D voxel mapping+GPU MPS)}: The proposed GPU-accelerated framework.
\end{enumerate}

\subsection{Performance comparison}
We report performance metrics: (i) processing time (ms), and (ii) plane-level intersection over union (IoU).
Table \ref{table:processing_time} summarizes the performance results for each method.

\begin{table}[h!]
  \centering
  \caption{Performance comparison across each frameworks}
  \label{table:processing_time}
  \begin{tabular}{l|c|c}
    \hline
    \multicolumn{1}{c|}{\textbf{Framework}} & \multirow{1}{*}{\textbf{Avg. time [ms]}} & \multirow{2}{*}{\textbf{IoU [\%]}} \\
    \multicolumn{1}{c|}{GPU mapping+MPS} & Mapping+MPS & \\
    \hline
    \hline
    1) Height map+CPU & \textbf{4.38}+1680 & 75.3 \\
    2) 3D voxel+CPU   & 6.07+3080 & \textbf{98.2} \\
    3) 3D voxel+GPU   & 6.07+\textbf{12.7} & \textbf{98.3} \\
    \hline
  \end{tabular}
\end{table}

\begin{figure}[h!]
  \centering
  \includegraphics[width=0.70\columnwidth]{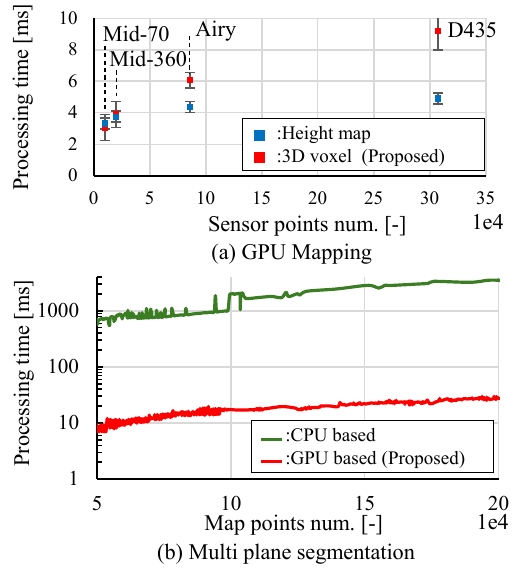}
  \caption{
    Processing time analysis in the five-step stair environment:
    (a) Mapping module processing time comparison across different sensor configurations. The proposed GPU-based 3D voxel mapping method achieves sufficiently high-speed mapping execution.
    (b) Multi-plane segmentation module processing time comparison. The proposed GPU-based multi-plane segmentation method maintains significantly superior performance compared to existing CPU-based methods.
  }
  \label{fig:sim_performance}
\end{figure}

\subsubsection{Processing time evaluation in multi-plane environment}
To quantify the processing time performance in multi-plane environments, we conducted comprehensive measurements in the five-step stair environment (Fig.~\ref{fig:sim_env}(a)). 
The processing time evaluation presented in Table~\ref{table:processing_time} was conducted using Airy LiDAR point cloud data, which enables accumulation of the most extensive range of map points among the tested sensor configurations.  The results show that the proposed framework achieves a processing time of 18.8\,ms in average, demonstrating a significant improvement over the height map-based method and the hybrid framework. 

Fig. \ref{fig:sim_performance}(a) illustrates the relationship between processing time and sensor point count for the mapping module across different sensor configurations.  In the GPU mapping module comparison, our implemented GPU-based 3D voxel mapping method requires marginally more computational time compared to the height map approach. This is attributed to the increased processing load in 3D space as sensor points increase, causing processing time to scale sensitively with point count. However, our 3D voxel mapping module achieves sufficiently high-speed mapping execution across diverse sensor configurations.

Fig.~\ref{fig:sim_performance}(b) presents a comparison of processing times between conventional CPU-based 3D point cloud segmentation methods and the proposed GPU-based multi-plane segmentation approach. While the processing time of the proposed method gradually increases with the number of accumulated map points, it consistently maintains significantly superior performance compared to existing CPU-based methods. The maximum processing time is 29.6\,ms, demonstrating the capability for operation at over 30\,Hz.

\subsubsection{Plane IoU evaluation in the single stage environment}
\begin{figure}[h]
  \centering
  \includegraphics[width=0.90\columnwidth]{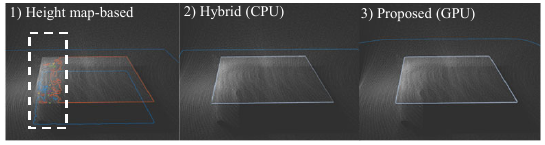}
  \caption{
    Comparative analysis of plane detection in the single-stage environment.
(a) Height map-based method cannot simultaneously represent top and bottom planes, reducing edge accuracy.
(b) Hybrid (CPU MPS) achieves accurate edge detection via 3D voxel accumulation.
(c) Proposed (GPU MPS) attains similar precision while maintaining real-time performance.
  }
  \label{fig:sim_results_step}
\end{figure}

Fig. \ref{fig:sim_results_step} shows plane detection performance in the single-stage environment, with quantitative metrics in Table \ref{table:processing_time}. The height map-based method has inherent limitations in representing both the stage top and ground plane simultaneously, resulting in degraded accuracy, especially at edges where the IoU is limited to 75.3\%. In contrast, the 3D voxel map-based approach accurately detects both planes. The proposed GPU-based method further achieves accuracy comparable to CPU-based approaches.

These results demonstrate that the proposed framework achieves both high-speed performance and high plane-level IoU, making it well-suited for legged robot locomotion applications.
Throughout the above experiments, the system operated stably on the Jetson AGX Orin, with peak memory usage measured at 16.8\,GB (26.3\%), demonstrating reliable performance.

\subsection{Contribution of cluster-level parallel processing}
\begin{figure}[h!]
  \centering
  \includegraphics[width=0.80\columnwidth]{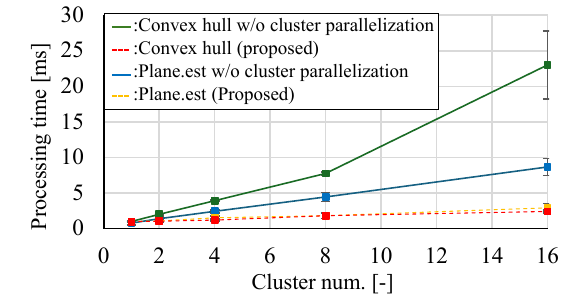}
  \caption{
    Processing time comparison with and without cluster-level parallelization. The proposed method with cluster-level parallelization consistently reduces computation time more efficiently as the number of clusters increases, demonstrating superior scalability.
  }
  \label{fig:sim_ablation}
\end{figure}

We evaluate the effect of cluster-level parallelization of RANSAC and convex hull, introduced to enable fast segmentation of multiple planar clusters containing a large number of points. 
In the evaluation, we measured processing time on multiple clusters (1, 2, 4, 8, and 16 clusters).   For each trial, we randomly selected the number of points \(M\) (10,000–30,000), and each cluster in that trial contained exactly \(M\) randomly generated points.
As a comparison, we considered the proposed method with cluster-level parallelization and a baseline method in which only a single cluster is processed in parallel on the GPU for sampling, iteration, and inlier evaluation (w/o cluster parallelization). Each condition was executed 1000 times, and the average processing time was reported. The results are shown in Fig.~\ref{fig:sim_ablation}. The proposed method consistently reduced computation time more efficiently as the number of clusters increased, demonstrating superior scalability.

\subsection{Application to legged robot locomotion}
We integrated the proposed method into a complete locomotion stack and evaluated its detection and planning capabilities in both simulation and on physical robots. Experiments were conducted using a wheel-legged robot equipped with Mid-360 and Mid-70 sensors, controlled via a Control Barrier Function (CBF)-based safe locomotion framework~\cite{t3_perceptive2024}, as well as the Go2\footnote{\url{https://www.unitree.com/go2}} quadruped robot outfitted with RoboSense Airy sensors. 
For self-localization in the real robot experiments, we employed LiDAR-based localization \cite{xu2021fast}.
All processing was performed onboard a Jetson AGX Orin.

\subsubsection{Multi-level plane segmentation in open-tread stair}
\begin{figure}[h]
  \centering
  \includegraphics[width=0.55\columnwidth]{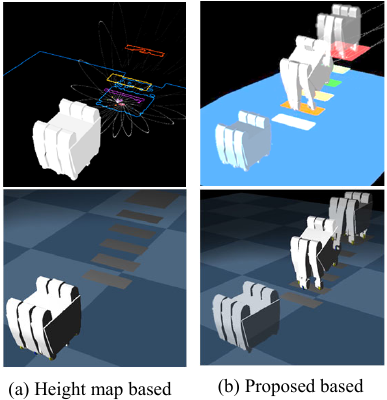}
  \caption{
  Locomotion experiment on open-tread stairs: (a) Height map-based method fails to detect multiple planes at identical (x, y) coordinates and cannot accomplish the task. (b) Proposed method detects multiple planes in 3D, enabling safe and effective locomotion.
  }
  \label{fig:stair_no_side}
\end{figure}

Fig.~\ref{fig:stair_no_side} illustrates the results of locomotion experiments on open-tread stairs composed solely of treads, such as Fig.~\ref{fig:proposed_demo}(b). The conventional height map-based method fails to accurately detect multi-level planes, resulting in locomotion failure. In contrast, the proposed method successfully recognizes multiple 3D planes and demonstrates effective locomotion capabilities in such challenging environments.

\subsubsection{Locomotion in Confined Spaces with Overhead Structures}
\begin{figure}[t]
  \centering
  \includegraphics[width=0.80\columnwidth]{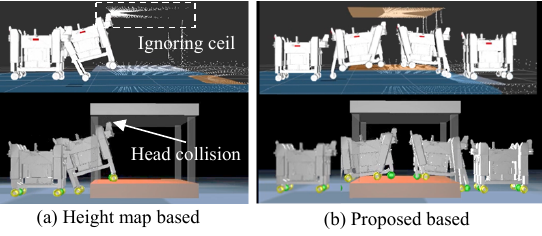}
  \caption{
    Comparison of locomotion in confined 3D spaces with overhead structures:
(a) Height-map-based method ignores overhead structures, causing collisions.
(b) Proposed 3D voxel-based method detects overhead planes and enables safe, collision-free planning.
  }
  \label{fig:multi_level_structured}
\end{figure}

As shown in Fig.~\ref{fig:multi_level_structured}, in narrow environments with overhead structures such as ceilings and beams, the proposed method performs real-time 3D multi-plane segmentation around the robot and supplies the resulting planar polygons to a barrier-function-based motion planner, achieving collision-free trajectory generation with respect to ceiling structures. In contrast, height map based planning cannot represent multiple z values at the same (x,y) and thus ignore overhead structures, making collisions likely during planning. These results indicate that a locomotion system integrating the proposed method enables safe navigation in complex 3D environments that include overhead structures.

\subsubsection{Real-time Detection of Small Objects and Avoidance/Overstepping}
\begin{figure}[h]
  \centering
  \includegraphics[width=0.80\columnwidth]{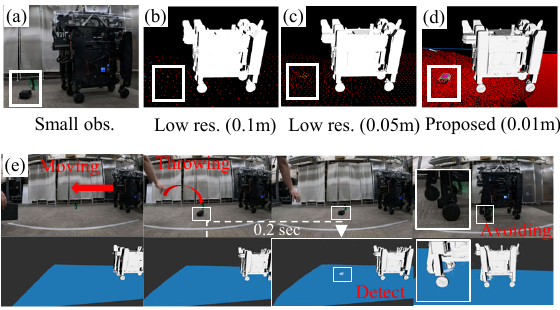}
  \caption{
    (a)-(d)Effect of voxel resolution on small-object representation.  Low-resolution maps make small objects ($0.07{\times}0.10{\times}0.08\,\mathrm{m}$) sparse and noise-like, whereas proposed high-resolution maps enable robust detection.
    (e) Real-time detection and overstepping of small obstacles with the wheel-legged robot with Mid-70 LiDAR.
  }
  \label{fig:real_comp_small_obs}
\end{figure}

Fig.~\ref{fig:real_comp_small_obs}(a)--(d) compares accumulated point clouds in 3D voxel maps at different resolutions. 
With low-resolution voxels, small obstacles remain extremely sparse and resemble noise, which hinders reliable plane or obstacle extraction. 
In contrast, the proposed high-resolution accumulation yields sufficient point support to recover fine-grained geometry, enabling robust detection and polygonization of small obstacles and other detailed structures that are critical for locomotion.

Fig.~\ref{fig:real_comp_small_obs}(e) demonstrates real-time detection and traversal of small obstacles with the wheel-legged robot. 
The proposed pipeline detects obstacles with an end-to-end latency of approximately $0.2$\,s and immediately feeds the resulting polygons to the planner, enabling both avoidance and precise overstepping. 
Beyond collision avoidance, this capability supports the detection and selection of small footholds, facilitating accurate foot placement in cluttered 3D environments.

\subsubsection{Safety-Enhanced Locomotion with Reinforcement Learning}
\begin{figure}[h]
  \centering
  \includegraphics[width=0.80\columnwidth]{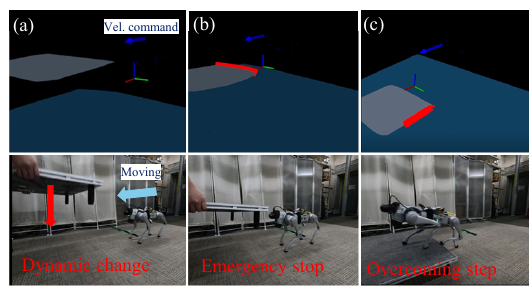}
  \caption{
   Safety-enhanced locomotion system on Go2 using RL policy with edge-based velocity filtering:
(a) Dynamic plane height change during robot locomotion, (b) Real-time change detection with emergency stop function, (c) Normal locomotion continuation for traversable obstacles. 
  }
  \label{fig:safety_rl}
\end{figure}

The 3D voxel map representation utilized in this study can be converted into a height map in the robot-centric coordinate system and directly employed as an input to reinforcement learning (RL) policies such as ~\cite{miki2022learning}. Moreover, by filtering the velocity commands used by the RL policy according to the distance to and the height difference from the planar boundaries (edges) extracted by the proposed method, safe motion generation can be achieved. The experimental videos demonstrate that, even in challenging scenarios such as dynamic 3D plane obstacles that are difficult to reproduce during training (Fig.~\ref{fig:safety_rl}) and falls caused by lateral approaches to staircases with unknown widths during training, the proposed edge-based velocity filtering successfully prevents failures through plane detection.

\begin{figure}[h]
  \centering
  \includegraphics[width=0.80\columnwidth]{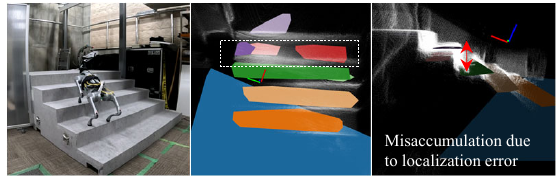}
  \caption{
    Failure case in plane segmentation during stair climbing.
  }
  \label{fig:real_failure_detect}
\end{figure}

Fig.~\ref{fig:real_failure_detect} illustrates a failure case in plane detection during stair climbing. When the robot's self-localization suddenly deviates significantly due to missteps or when drift errors accumulate over time, the aggregation of 3D point clouds becomes distorted, leading to a decline in plane detection accuracy. 
Robust handling of such self-localization errors remains an unresolved challenge. As future directions, incorporating post-processing drift correction of detected polygons as in \cite{polygon_mapping2024}, or integrating the proposed approach into SLAM frameworks such as \cite{hoss2024covoxslam}, are considered promising solutions.

\subsubsection{Operation with diverse real robots and sensors}
To verify the versatility of the proposed method, we conducted experiments using different sensor configurations. For detailed, please refer to our github page.

\section{CONCLUSTION}
This paper presents a real-time multi-plane segmentation method based on GPU-accelerated high-resolution 3D voxel mapping for legged robot locomotion. The proposed framework generates precise 3D polygonal representations in real time, accurately detects multi-layered planar surfaces in complex environments, and enables safe locomotion such as stair traversal and navigation beneath overhangs. Experimental validation demonstrates that the method achieves real-time performance with high accuracy in both simulation and real-world robot platforms. The system attains 3D voxel mapping with 0.01 m resolution on Jetson AGX Orin hardware and effectively applies the detected planar information to locomotion control in challenging 3D environments.

Several limitations remain in this study. First, 3D environment representation requires much larger memory (16.8 GB in evaluation) than height maps, inherently restricting the mapping range. Addressing this issue calls for more efficient data structures for GPU-based point cloud accumulation and improved neighborhood search methods, both of which remain important challenges. Second, while the proposed method effectively leverages planar information for locomotion in structured environments, its application to unstructured terrains such as outdoor off-road settings remains difficult. As a future direction, extending the framework to provide simplified representations not only for traversable but also for obstacle point clouds could broaden its use in navigation tasks such as path planning and collision avoidance.

\bibliographystyle{IEEEtran}
\bibliography{ref}

\begin{thebibliography}{10}
\providecommand{\url}[1]{#1}
\csname url@rmstyle\endcsname
\providecommand{\newblock}{\relax}
\providecommand{\bibinfo}[2]{#2}
\providecommand\BIBentrySTDinterwordspacing{\spaceskip=0pt\relax}
\providecommand\BIBentryALTinterwordstretchfactor{4}
\providecommand\BIBentryALTinterwordspacing{\spaceskip=\fontdimen2\font plus
\BIBentryALTinterwordstretchfactor\fontdimen3\font minus
  \fontdimen4\font\relax}
\providecommand\BIBforeignlanguage[2]{{%
\expandafter\ifx\csname l@#1\endcsname\relax
\typeout{** WARNING: IEEEtran.bst: No hyphenation pattern has been}%
\typeout{** loaded for the language `#1'. Using the pattern for}%
\typeout{** the default language instead.}%
\else
\language=\csname l@#1\endcsname
\fi
#2}}

\bibitem{bhatti2015survey}
J.~Bhatti, A.~R. Plummer, P.~Iravani, and B.~Ding, ``A survey of dynamic robot
  legged locomotion,'' in \emph{Proc. Int. Conf. Fluid Power and Mechatronics
  (FPM)}, 2015, pp. 770--775.

\bibitem{yoshiike2017development}
T.~Yoshiike, M.~Kuroda, R.~Ujino, H.~Kaneko, H.~Higuchi, S.~Iwasaki,
  Y.~Kanemoto, M.~Asatani, and T.~Koshiishi, ``Development of experimental
  legged robot for inspection and disaster response in plants,'' in \emph{Proc.
  IEEE/RSJ Int. Conf. Intell. Robots Syst. (IROS)}, 2017, pp. 4869--4876.

\bibitem{gehring2021anymal}
C.~Gehring, P.~Fankhauser, L.~Isler, R.~Diethelm, S.~Bachmann, M.~Potz,
  L.~Gerstenberg, and M.~Hutter, ``Anymal in the field: Solving industrial
  inspection of an offshore hvdc platform with a quadrupedal robot,'' in
  \emph{Field and Service Robotics: Results of the 12th International
  Conference}, 2021, pp. 247--260.

\bibitem{chen2024autonomous}
Y.~Chen, Z.~Wei, S.~G. Vougioukas, S.~K. Gupta, and Q.~Nguyen, ``Autonomous
  visual navigation for quadruped robot in farm operation,'' in \emph{Proc.
  IEEE Int. Conf. Autom. Sci. Eng. (CASE)}, 2024, pp. 3518--3524.

\bibitem{grandia2021multi}
R.~Grandia, A.~J. Taylor, A.~D. Ames, and M.~Hutter, ``Multi-layered safety for
  legged robots via control barrier functions and model predictive control,''
  in \emph{Proc. IEEE Int. Conf. Robot. Autom. (ICRA)}, 2021, pp. 8352--8358.

\bibitem{sato2022robust}
S.~Sato, Y.~Kojio, Y.~Kakiuchi, K.~Kojima, K.~Okada, and M.~Inaba, ``Robust
  humanoid walking system considering recognized terrain and robots' balance,''
  in \emph{Proc. IEEE/RSJ Int. Conf. Intell. Robots Syst. (IROS)}, 2022, pp.
  8298--8305.

\bibitem{perceptive_locomotion2023}
R.~Grandia, F.~Jenelten, S.~Yang, F.~Farshidian, and M.~Hutter, ``Perceptive
  locomotion through nonlinear model-predictive control,'' \emph{IEEE Trans.
  Robot.}, vol.~39, no.~5, pp. 3402--3421, 2023.

\bibitem{elevation2018}
P.~Fankhauser, M.~Bloesch, and M.~Hutter, ``Probabilistic terrain mapping for
  mobile robots with uncertain localization,'' \emph{IEEE Robot. Autom. Lett.},
  vol.~3, no.~4, pp. 3019--3026, 2018.

\bibitem{elevation_gpu2022}
T.~Miki, L.~Wellhausen, R.~Grandia, F.~Jenelten, T.~Homberger, and M.~Hutter,
  ``Elevation mapping for locomotion and navigation using {GPU},'' in
  \emph{Proc. IEEE/RSJ Int. Conf. Intell. Robots Syst. (IROS)}, 2022, pp.
  2273--2280.

\bibitem{mem2023}
G.~Erni, J.~Frey, T.~Miki, M.~Mattamala, and M.~Hutter, ``Mem: Multi-modal
  elevation mapping for robotics and learning,'' in \emph{Proc. IEEE/RSJ Int.
  Conf. Intell. Robots Syst. (IROS)}, 2023, pp. 11\,011--11\,018.

\bibitem{supervoxel_plane2017}
T.~Zhang, S.~Caron, and Y.~Nakamura, ``Supervoxel plane segmentation and
  multi-contact motion generation for humanoid stair climbing,'' \emph{Int. J.
  Humanoid Robot.}, vol.~14, no.~01, p. 1650022, 2017.

\bibitem{bertrand2020detecting}
S.~Bertrand, I.~Lee, B.~Mishra, D.~Calvert, J.~Pratt, and R.~Griffin,
  ``Detecting usable planar regions for legged robot locomotion,'' in
  \emph{Proc. IEEE/RSJ Int. Conf. Intell. Robots Syst. (IROS)}, 2020, pp.
  4736--4742.

\bibitem{deits2014}
R.~Deits and R.~Tedrake, ``Footstep planning on uneven terrain with
  mixed-integer convex optimization,'' in \emph{Proc. IEEE-RAS Int. Conf.
  Humanoid Robots (Humanoids)}, 2014, pp. 279--286.

\bibitem{tonneau2018}
S.~Tonneau, A.~D. Prete, J.~Pettre, C.~Park, D.~Manocha, and N.~Mansard, ``An
  efficient acyclic contact planner for multiped robots,'' \emph{IEEE Trans.
  Robot.}, vol.~34, no.~3, pp. 586--601, 2018.

\bibitem{grilli2017review}
E.~Grilli, F.~Menna, and F.~Remondino, ``A review of point clouds segmentation
  and classification algorithms,'' \emph{Int. Arch. Photogramm. Remote Sens.
  Spatial Inf. Sci.}, vol.~42, pp. 339--344, 2017.

\bibitem{sun2024review}
Y.~Sun, X.~Zhang, and Y.~Miao, ``A review of point cloud segmentation for
  understanding 3d indoor scenes,'' \emph{Visual Intelligence}, vol.~2, no.~1,
  p.~14, 2024.

\bibitem{miki2022learning}
T.~Miki, J.~Lee, J.~Hwangbo, L.~Wellhausen, V.~Koltun, and M.~Hutter,
  ``Learning robust perceptive locomotion for quadrupedal robots in the wild,''
  \emph{Sci. Robot.}, vol.~7, no.~62, p. eabk2822, 2022.

\bibitem{xu2024dexterous}
Z.~Xu, A.~H. Raj, X.~Xiao, and P.~Stone, ``Dexterous legged locomotion in
  confined 3d spaces with reinforcement learning,'' in \emph{Proc. IEEE Int.
  Conf. Robot. Autom. (ICRA)}, 2024, pp. 11\,474--11\,480.

\bibitem{kovalev2023combining}
V.~Kovalev, A.~Shkromada, H.~Ouerdane, and P.~Osinenko, ``Combining
  model-predictive control and predictive reinforcement learning for stable
  quadrupedal robot locomotion,'' \emph{arXiv preprint arXiv:2307.07752}, 2023.

\bibitem{yang2023continuous}
Y.~Yang, X.~Meng, W.~Yu, T.~Zhang, J.~Tan, and B.~Boots, ``Continuous versatile
  jumping using learned action residuals,'' in \emph{Proc. Learn. Dyn. Control
  (L4DC)}, 2023, pp. 770--782.

\bibitem{yang2024agile}
Y.~Yang, G.~Shi, C.~Lin, X.~Meng, R.~Scalise, M.~G. Castro, W.~Yu, T.~Zhang,
  D.~Zhao, J.~Tan, \emph{et~al.}, ``Agile continuous jumping in discontinuous
  terrains,'' in \emph{2025 IEEE International Conference on Robotics and
  Automation (ICRA)}.\hskip 1em plus 0.5em minus 0.4em\relax IEEE, 2025, pp.
  10\,245--10\,252.

\bibitem{tan2025planarsplatting}
B.~Tan, R.~Yu, Y.~Shen, and N.~Xue, ``Planarsplatting: Accurate planar surface
  reconstruction in 3 minutes,'' in \emph{Proc. IEEE/CVF Conf. Comput. Vis.
  Pattern Recognit. (CVPR)}, 2025, pp. 1190--1199.

\bibitem{polytopic2022}
Z.~Xu, H.~Zhu, H.~Chen, and W.~Zhang, ``Polytopic planar region
  characterization of rough terrains for legged locomotion,'' in \emph{Proc.
  IEEE/RSJ Int. Conf. Intell. Robots Syst. (IROS)}, 2022, pp. 8682--8689.

\bibitem{planar_gpu2021}
B.~Mishra, D.~Calvert, S.~Bertrand, S.~McCrory, R.~Griffin, and H.~E. Sevil,
  ``{GPU}-accelerated rapid planar region extraction for dynamic behaviors on
  legged robots,'' in \emph{Proc. IEEE/RSJ Int. Conf. Intell. Robots Syst.
  (IROS)}, 2021, pp. 8493--8499.

\bibitem{polygon_mapping2024}
T.~Bin, J.~Yao, T.~L. Lam, and T.~Zhang, ``Real-time polygonal semantic mapping
  for humanoid robot stair climbing,'' in \emph{Proc. IEEE-RAS Int. Conf.
  Humanoid Robots (Humanoids)}, 2024, pp. 866--873.

\bibitem{polylidar3d2020}
J.~Castagno and E.~Atkins, ``Polylidar3d: Fast polygon extraction from 3d
  data,'' \emph{Sensors}, vol.~20, no.~17, p. 4819, 2020.

\bibitem{liu2021low}
Z.~Liu, F.~Zhang, and X.~Hong, ``Low-cost retina-like robotic lidars based on
  incommensurable scanning,'' \emph{IEEE/ASME Trans. Mechatronics}, vol.~27,
  no.~1, pp. 58--68, 2021.

\bibitem{t3_perceptive2024}
N.~Takasugi, M.~Kinoshita, Y.~Kamikawa, R.~Tsuzaki, A.~Sakamoto, T.~Kai, and
  Y.~Kawanami, ``Real-time perceptive motion control using control barrier
  functions with analytical smoothing for six-wheeled-telescopic-legged robot
  tachyon 3,'' in \emph{Proc. IEEE/RSJ Int. Conf. Intell. Robots Syst. (IROS)},
  2024, pp. 6802--6809.

\bibitem{nvblox2024}
A.~Millane, H.~Oleynikova, E.~Wirbel, R.~Steiner, V.~Ramasamy, D.~Tingdahl, and
  R.~Siegwart, ``nvblox: {GPU}-accelerated incremental signed distance field
  mapping,'' in \emph{Proc. IEEE Int. Conf. Robot. Autom. (ICRA)}, 2024, pp.
  2698--2705.

\bibitem{hoss2024covoxslam}
E.~H{\"o}ss and P.~D. Crist{\'o}foris, ``covoxslam: Gpu accelerated globally
  consistent dense slam,'' in \emph{Proc. IEEE Int. Conf. Robot. Autom.
  (ICRA)}.\hskip 1em plus 0.5em minus 0.4em\relax IEEE, 2025, pp. 6198--6205.

\bibitem{overbye2022g}
T.~Overbye and S.~Saripalli, ``G-{VOM}: A {GPU} accelerated voxel off-road
  mapping system,'' in \emph{Proc. IEEE Intell. Vehicles Symp. (IV)}, 2022, pp.
  1480--1486.

\bibitem{stepanas2022ohm}
K.~Stepanas, J.~Williams, E.~Hern{\'a}ndez, F.~Ruetz, and T.~Hines, ``Ohm:
  {GPU}-based occupancy map generation,'' \emph{IEEE Robot. Autom. Lett.},
  vol.~7, no.~4, pp. 11\,078--11\,085, 2022.

\bibitem{golub2013matrix}
G.~H. Golub and C.~F.~V. Loan, \emph{Matrix Computations}, 4th~ed.\hskip 1em
  plus 0.5em minus 0.4em\relax Johns Hopkins University Press, 2013.

\bibitem{gpu_euclidean2020}
A.~Nguyen, A.~M. Cano, M.~Edahiro, and S.~Kato, ``Fast euclidean cluster
  extraction using {GPUs},'' \emph{J. Robotics Mechatronics}, vol.~32, no.~3,
  pp. 548--560, 2020.

\bibitem{gpu_convex_hull_2022}
A.~Keith, H.~Ferrada, and C.~A. Navarro, ``Accelerating the convex hull
  computation with a parallel {GPU} algorithm,'' in \emph{Proc. Int. Conf.
  Chilean Comput. Sci. Soc. (SCCC)}, 2022, pp. 1--7.

\bibitem{todorov2012mujoco}
E.~Todorov, T.~Erez, and Y.~Tassa, ``Mujoco: A physics engine for model-based
  control,'' in \emph{Proc. IEEE/RSJ Int. Conf. Intell. Robots Syst. (IROS)},
  2012, pp. 5026--5033.

\bibitem{xu2021fast}
W.~Xu and F.~Zhang, ``Fast-lio: A fast, robust lidar-inertial odometry package
  by tightly-coupled iterated kalman filter,'' \emph{IEEE Robot. Autom. Lett.},
  vol.~6, no.~2, pp. 3317--3324, 2021.

\end{thebibliography}

\end{document}